\pdfoutput=1

\documentclass[11pt]{article}

\usepackage{acl_template/acl}

\usepackage{times}
\usepackage{latexsym}

\usepackage[T1]{fontenc}

\usepackage[utf8]{inputenc}

\usepackage{microtype}

\usepackage{graphicx}
\usepackage{xspace}

\usepackage{epsfig}
\usepackage{graphicx}
\usepackage{wrapfig}
\usepackage{subcaption}

\usepackage{amsmath, amsthm, amssymb}
\usepackage{textcomp}
\usepackage{stmaryrd}
\usepackage{upgreek}
\usepackage{bm}
\usepackage{cases}
\usepackage{mathtools}
\usepackage{arydshln}
\usepackage{multirow}

\usepackage[ruled,vlined]{algorithm2e}

\usepackage{paralist}

\usepackage{cite}

\usepackage{multirow}
\usepackage{rotating}
\usepackage{booktabs}

\usepackage{url}
\usepackage{xspace}
\usepackage[T1]{fontenc}
\usepackage{pifont} 

\usepackage{pdfrender}
\newcommand*{\boldcheckmark}{%
  \textpdfrender{
    TextRenderingMode=FillStroke,
    LineWidth=.5pt, 
  }{\checkmark}%
}

\newcommand{\visitron}[0]{\textsc{Visitron}\xspace}
\newcommand{\visitronBold}[0]{\textbf{\textsc{Visitron}}\xspace}

\newcommand{\navigator}[0]{\textsc{Navigator}\xspace}

\newcommand{\guide}[0]{\textsc{Guide}\xspace}

\newcommand{\oracle}[0]{\textsc{Oracle}\xspace}
\newcommand{\mixed}[0]{\textsc{Mixed}\xspace}

\newcommand{\questioner}[0]{\textsc{Questioner}\xspace}

\newcommand{\oscar}[0]{\textsc{Oscar}\xspace}

\newcommand{\rmm}[0]{\textsc{Rmm}\xspace}

\newcommand{\stoptoken}[0]{\texttt{STOP}\xspace}
\newcommand{\forwardtoken}[0]{\texttt{FORWARD}\xspace}
\newcommand{\lefttoken}[0]{\texttt{LEFT}\xspace}
\newcommand{\righttoken}[0]{\texttt{RIGHT}\xspace}

\newcommand{\GP}[0]{\textbf{\texttt{GP}}\xspace}
\newcommand{\SPL}[0]{\textbf{\texttt{SPL}}\xspace}
\newcommand{\SR}[0]{\textbf{\texttt{SR}}\xspace}
\newcommand{\nDTW}[0]{\textbf{\texttt{nDTW}}\xspace}

\newcommand{\abbrvcoloring}[1]{\textcolor{magenta}{#1}}

\newcommand{\cmark}{\ding{51}}%
\newcommand{\xmark}{\ding{55}}%

\newcommand{\dia}[0]{dialogue\xspace}

\newcommand{\clsToken}[0]{\mathtt{[CLS]}\xspace}
\newcommand{\sepToken}[0]{\mathtt{[SEP]}\xspace}
\newcommand{\tarToken}[0]{\mathtt{[TAR]}\xspace}
\newcommand{\navToken}[0]{\mathtt{[NAV]}\xspace}
\newcommand{\guideToken}[0]{\mathtt{[GUI]}\xspace}
\newcommand{\maskToken}[0]{\mathtt{[MASK]}\xspace}

\definecolor{red}{RGB}{128, 0, 0}

\DeclareMathAlphabet\mathbfcal{OMS}{cmsy}{b}{n}

\usepackage{amsmath,amsfonts,bm}

\def\eqref#1{equation~\ref{#1}}

\def\1{\bm{1}}

\def\vtheta{{\bm{\theta}}}
\def\vtau{{\bm{\tau}}}
\def\va{{\bm{a}}}

\def\vh{{\bm{h}}}

\def\vq{{\bm{q}}}

\def\vs{{\bm{s}}}

\def\vv{{\bm{v}}}
\def\vw{{\bm{w}}}
\def\vx{{\bm{x}}}

\def\vz{{\bm{z}}}

\DeclareMathAlphabet{\mathsfit}{\encodingdefault}{\sfdefault}{m}{sl}
\SetMathAlphabet{\mathsfit}{bold}{\encodingdefault}{\sfdefault}{bx}{n}

\def\gA{{\mathcal{A}}}

\def\gD{{\mathcal{D}}}

\def\gL{{\mathcal{L}}}
\def\gM{{\mathcal{M}}}

\def\gS{{\mathcal{S}}}

\def\sR{{\mathbb{R}}}

\newcommand{\E}{\mathbb{E}}

\usepackage{verbatim}

\hypersetup{
    pagebackref=true,breaklinks=true,colorlinks,bookmarks=false
    }

\title{\abbrvcoloring{VISITRON}: \abbrvcoloring{Vi}sual \abbrvcoloring{S}emantics-Aligned \abbrvcoloring{I}nteractively \abbrvcoloring{Tr}ained \abbrvcoloring{O}bject-\abbrvcoloring{N}avigator}

\author{Ayush Shrivastava$^1$\Thanks{ Work done as an intern at Amazon Alexa AI. Code available at: \url{www.github.com/alexa/visitron}}, Karthik Gopalakrishnan$^2$, Yang Liu$^2$, Robinson Piramuthu$^2$,\\
{\bf Gokhan T\"{u}r$^2$, Devi Parikh$^1$, Dilek Hakkani-T\"{u}r$^2$}\\
$^1$Georgia Tech, $^2$Amazon Alexa AI\\
{\tt\small \{ayshrv, parikh\}@gatech.edu}\\ {\tt\small \{karthgop, yangliud, robinpir, gokhatur, hakkanit\}@amazon.com}
}

\begin{document}
\maketitle

\begin{abstract}
Interactive robots navigating photo-realistic environments need to be trained to effectively leverage and handle the dynamic nature of \dia
in addition to the challenges underlying vision-and-language navigation (VLN).
In this paper, we present \visitron, a multi-modal Transformer-based navigator better suited to the interactive regime inherent to Cooperative Vision-and-Dialog Navigation (CVDN).
\visitron is trained to: i) identify and associate object-level concepts and semantics between the environment and \dia history, ii) identify when to interact vs. navigate via imitation learning of a binary classification head. We perform extensive pre-training and fine-tuning ablations with \visitron to gain empirical insights and improve performance on CVDN. \visitron's ability to identify when to interact leads to a natural generalization of the game-play mode introduced by \newcite{roman2020rmm} for enabling the use of such models in different environments.
\visitron is competitive with models on the static CVDN leaderboard and attains state-of-the-art performance on the Success weighted by Path Length (SPL) metric.
\end{abstract}

\section{Introduction}\label{intro}

Large pre-trained Transformer-based language models~\citep{vaswani2017attention} are ubiquitous in natural language processing (NLP) and have performed very well in interactive settings such as open-domain~\citep{gopalakrishnan2019topical, huang2020challenges} and task-oriented \dia~\citep{kim2020beyond}. 
The success of Transformers and the pre-train/fine-tune paradigm in NLP has also inspired their adoption in vision-and-language research, with cross-modal representations being learned~\citep{li2020oscar} and utilized towards tasks like image and object captioning, visual question answering, visual commonsense reasoning and visual \dia.

\begin{figure}[t]
    \centering
    \vspace{-5pt}
    \includegraphics[width=0.97\linewidth]{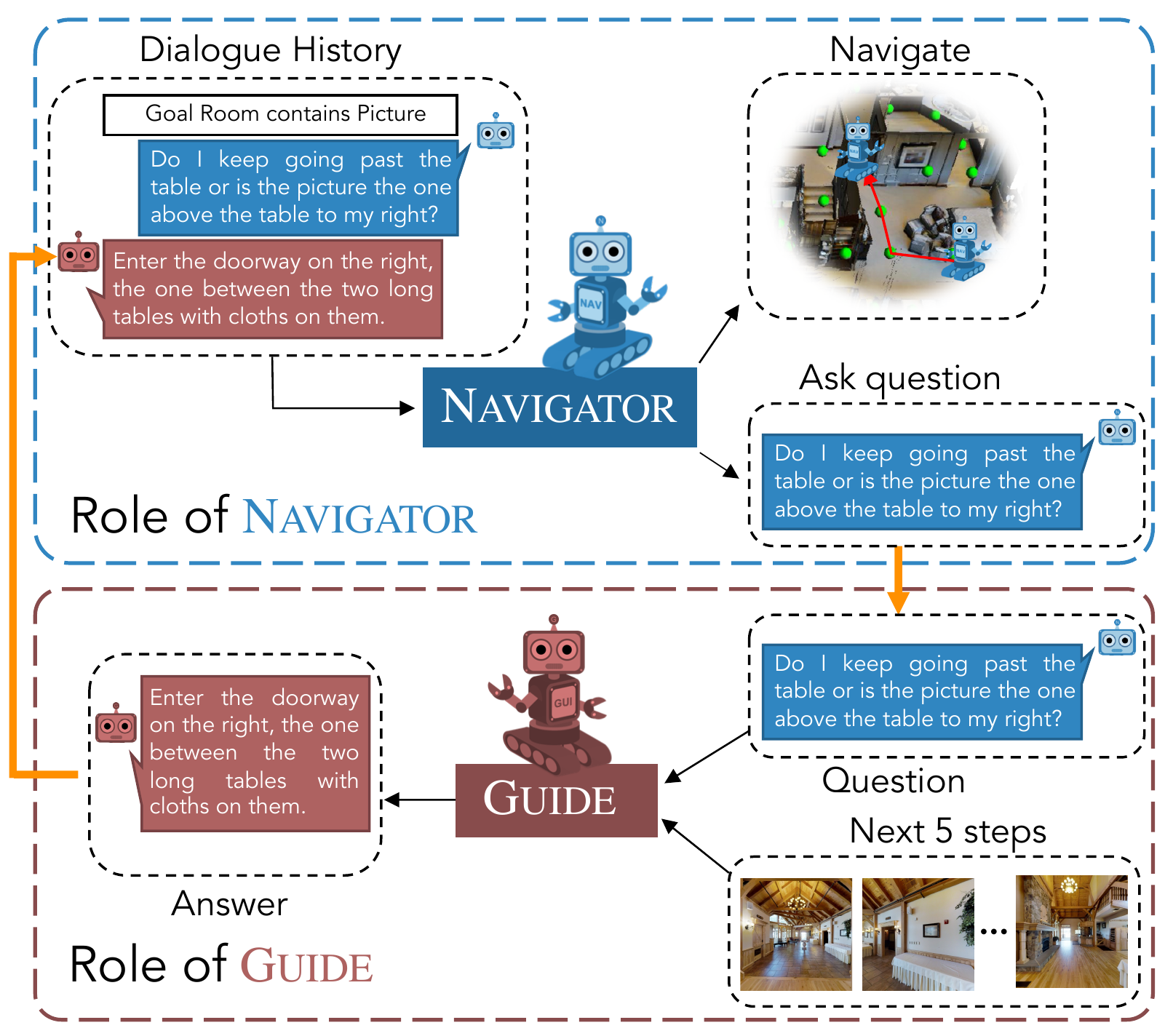}
    \caption{Cooperative Vision-and-Dialog Navigation (CVDN) with Dynamic Question-Asking}
    \vspace{-15pt}
    \label{fig:visualization}
\end{figure}

Vision-and-language navigation (VLN) is a challenging cross-modal research task in which agents need to learn to navigate in response to natural language instructions in simulated photo-realistic environments. VLN has been studied extensively with the advent of the Room-to-Room (R2R) dataset~\citep{anderson2018vision} and there has been growing interest recently in pushing the pre-train/fine-tune paradigm towards VLN, with work on leveraging disembodied corpora~\citep{majumdar2020improving} to learn cross-modal pre-trained representations that can improve embodied VLN performance.
As depicted in Figure~\ref{fig:visualization}, the Cooperative Vision-and-Dialog Navigation (CVDN) dataset~\citep{thomason2020vision} allows for \dia with a guide during navigation: a navigator can ask natural language questions to a guide when it needs assistance and the guide responds in natural language by using privileged knowledge of the environment accessible only to it, thus expanding beyond the traditional VLN task towards deployable interactive agents that are more robust and generalizable. But preliminary navigator modeling using CVDN is still VLN-style via the Navigation from Dialog History (NDH) task, treating the \dia history as a static instruction.

In this paper, we present work on training \visitron, a multi-modal Transformer-based navigator with a focus on tackling challenges unique to CVDN: i) moving beyond rote memorization to associative learning in order to learn to identify and acquire visio-linguistic concepts and semantics while interacting in new environments, and ii) learning when to ask questions~\citep{chi2020just}. \visitron builds off the recent cross-modal object-semantics aligned pre-training (\oscar) strategy and uses object-tags as explicit anchor points during training to learn to associate the environment's visual semantics with the textual \dia history, thus allowing for interaction/experience-grounded~\citep{bisk-etal-2020-experience} visio-linguistic concepts and semantics identification and acquisition. \visitron is trained in a data-driven fashion to identify when to engage in \dia, i.e., ask questions, vs. when to navigate, thus providing the first known empirical baselines for this task. We also present empirical results from various first-principles modeling ablations performed with \visitron. We demonstrate that for CVDN, panoramic viewpoint selection is a better formulation than discrete turn-based action prediction, akin to what has been seen on VLN with R2R~\citep{fried2018speaker}. We observe that multi-task learning with long-trajectory VLN datasets leads to significant CVDN performance gains relative to training on CVDN alone. \visitron is competitive with models on the leaderboard for the \textit{static} NDH task on EvalAI~\citep{yadav2019evalai}, attaining state-of-the-art performance on the Success weighted by Path Length (SPL) metric. Given \visitron's design and ability to identify when to engage in dialogue, we also propose a generalization of the game-play mode introduced by \newcite{roman2020rmm} for jointly fine-tuning and evaluating \visitron and future such models with pre-trained guides to help them easily adapt to their guides' capabilities.

\section{Background}\label{prelims}
\subsection{Vision-and-Language Navigation}\label{VLN}
The Vision-and-Language Navigation (VLN) task requires an agent spawned in an indoor environment at a starting position $\vs_0$ to follow natural language instructions $\vx$ and navigate to a target position $\vs_{goal}$. This can also be seen as a Partially Observable Markov Decision Process $\gM = \langle \gS, \gA, P_s, r \rangle$ where $\gS$ is the visual state space, $\gA$ is the discrete action space, $P_s$ is the unknown environment distribution from which the next state is drawn and $r \in \mathbb{R}$ is the reward function~\citep{hao2020towards}. At a given time step $t$, the agent receives an RGB image observation $obs(\vs_t)$, where $\vs_t \in \gS$. Based on the observation, the agent takes an action $\va_t \in \gA$, transitions into the next state $\vs_{t+1}$ drawn as follows: $\vs_{t+1} \sim P_{s}( \cdot | \vs_t, \va_t)$, and receives a new image observation $obs(\vs_{t+1})$. To end the episode, the agent must select the special \stoptoken action. A $T$-step trajectory can be represented as $\vtau = [ \vs_0, \va_0, \vs_1, \va_1, \ldots, \vs_T, \va_T]$. The episode is considered successful if the agent stops within $\epsilon$ distance of the goal, i.e., $|\vs_T - \vs_{goal}|\leq \epsilon$.
Using a training dataset $\gD = \{( \vtau, \vx) \}$ consisting of \textit{expert} trajectory $\vtau$ and instructions $\vx$ pairs, the goal is to train a policy $\pi_\vtheta (\vtau | \vx)$ with $\vtheta$ parameters that maximizes the log-likelihood of the target trajectory given instructions $\vx$:

\begin{equation}
\label{eq:max_likelihood}
\begin{split}
\max_{(\vtau, \vx)\sim\gD} \gL_{\vtheta}(\vtau, \vx) & = \log \pi_{\vtheta}(\vtau | \vx) \\
& = \sum_{t=0}^{T} \log \pi_{\vtheta} (\va_t | \vs_t, \vx)
\end{split}
\end{equation}

Several datasets have been released for VLN based on Matterport3D~\citep{Matterport3D}, a large-scale RGB-D dataset containing $\sim$10000 panoramic views from $\sim$194000 RGB-D images of 90 building-scale scenes. The most popular VLN dataset based on Matterport3D is the Room-to-Room (R2R) dataset~\citep{anderson2018vision}, containing $\sim$7200 trajectories and 3 natural language instructions per trajectory. For validation and test sets, \textit{seen} and \textit{unseen} splits are created to easily evaluate how well an agent generalizes. Room-4-Room (R4R)~\citep{jain2019stay} is an augmentation of R2R wherein existing short trajectories in R2R are joined to form longer, challenging trajectories. Room-across-Room (RxR)~\citep{ku2020room} is a newly introduced dataset with several properties, including but not limited to multilingual instructions, larger scale (for each language, $\sim$14000 trajectories with 3 instructions per trajectory), fine-grained spatio-temporal grounding and follower demonstrations.

A navigating agent's actions typically belong in a pre-defined discrete set comprising options such as \forwardtoken, \lefttoken, \righttoken, etc. Predicting the next best action from this low-level visuomotor space~\citep{fried2018speaker} of actions is referred to as \textit{turn-based action prediction}. Given the nature of the aforementioned VLN datasets, it is also possible to have a navigating agent's actions belong in the panoramic space, wherein the agent selects the next best viewpoint in the navigation graph from the panoramic space visible to it at its current location. This is referred to as \textit{viewpoint selection}.

\subsection{Cooperative Vision-and-Dialog Navigation} \label{CVDN}
Cooperative Vision-and-Dialog Navigation (CVDN) is a recently introduced dataset~\citep{thomason2020vision} collected by partnering crowd-workers in simulated photo-realistic environments. One worker acts as a \navigator, seeking to navigate to a goal and interacting in natural language with a \guide along the way if it needs assistance. The other worker acts as a \guide, answering the \navigator's questions while having privileged access to the best next steps the \navigator should take according to an \oracle full-state shortest path planner. The collection of each CVDN instance begins with the state $(\gS, T_O, \vs_0, G)$, where $\gS$ is the environment in which the agents are placed, $\vs_0$ is the start location of the \navigator, $G$ is the goal region and $T_O$ is the initial hint given to both agents about the goal region containing object $O$. At any time step $t$, the \navigator can make one of three choices: i) take a sequence of $k_t$ navigation steps $N_t = [n_t^{1}, n_t^{2}, \ldots, n_t^{k_t}]$, ii) ask a question $Q_t$ to the \guide, iii) declare its current position as the goal region. If a question is asked, the \guide looks at $l$ next steps along the shortest path to the goal
and replies with an answer $A_t$. The instance ends when the \navigator reaches $G$. Thus, a CVDN instance comprises $\big[ (\gS, T_O, \vs_o, G), \langle N_0, Q_1, A_1, N_1, Q_2, A_2, N_2, \ldots, \allowbreak\text{} Q_m, A_m, N_m \rangle \big] $, where $m$ is the number of \dia exchanges between the \navigator and \guide, and $N_0$ is the sequence of navigation steps before the $1^{\text{st}}$ exchange.

\subsubsection{Navigation from Dialog History (NDH)} \label{NDH}
With the CVDN dataset, the NDH task for the \navigator was introduced~\citep{thomason2020vision}, in which the \navigator needs to navigate towards a goal given a \dia history.
Specifically, the \navigator is spawned at the \textit{terminal} position of $N_{t-1}$ (or $\vs_0$ in the case of $N_0$) in environment $\gS$ and is given $(T_O, Q_{1:t}, A_{1:t})$. The task is to predict the navigation steps that bring the agent closer to the goal region $G$. To train a \navigator \textit{agent} for this task, the navigation steps needed for supervision from the dataset can be provided in any of the three forms: i) \textit{human} \navigator steps, $N_t$: the navigation steps that were taken by the \textit{human} \navigator after the \dia exchange at time step $t$, ii) \oracle steps, $O_t$: the shortest path steps accessible to the \guide when it gave the answer $A_t$, iii) \mixed: a mix of both \textit{human} \navigator and \oracle supervision where the supervision path is $N_t$ when $e(O_t) \in N_t$, and $O_t$ otherwise, where $e(\cdot)$ represents the \textit{terminal} position of a sequence of navigation steps. The \textit{agent} \navigator is trained VLN-style using Equation~\ref{eq:max_likelihood} on NDH instances extracted as described above from the CVDN instances, and evaluated on NDH instances using VLN metrics such as Goal Progress and Success weighted by Path Length (SPL), defined in Section~\ref{evalmetrics}. In the CVDN literature, it has been observed that \mixed supervision typically performs the best, followed by \oracle and \textit{human} \navigator supervision respectively. However, for the purposes of all our experiments, we pick the \textit{human} \navigator supervision mode to establish a lower-bound on performance for \visitron.

\subsubsection{Gameplay Mode}\label{CVDN-GM}
In the CVDN dataset, a \textit{human} \navigator cooperates with a \textit{human} \guide to find a goal region $G$ with target object $O$. \newcite{roman2020rmm} introduced the game-play mode, which is essentially an \textit{agent}-\textit{agent} replica of this dynamic dataset creation process wherein the two trained agents consume each other's outputs. This mode can be applied during both fine-tuning and evaluation and helps understand how well a pre-trained \navigator agent adapts to the capabilities of different \guide agents in a dynamic/interactive setting. For the sake of consistency with game-play mode notation introduced by \newcite{roman2020rmm}, we denote the role of asking questions that is intrinsic to the \navigator by \questioner. Thus, in a game-play mode episode, at $t=0$ (prior to the first QA exchange), the \navigator takes $N_0$ steps given the initial hint $T_O$. For time steps $t>0$, the \questioner generates a question $Q_t$, \guide generates an answer $A_t$ having access to the next $l$ steps in the shortest path, and then \navigator generates $N_t$ navigation steps of length $k_t$. All agents have access to the entire visual navigation ($N_{0:t-1}$) and \dia ($Q_{1:t-1}A_{1:t-1}$) histories in addition to the initial hint $T_O$. The \questioner asks questions every 4\textsuperscript{th} time-step, which is a hard-coded heuristic by \newcite{roman2020rmm} since their \navigator does not know \textit{when} to ask questions. The episode ends when the \navigator declares that the current position is in the goal region $G$ or a maximum number of turns (20) are played. \navigator's performance in game-play mode is measured using Goal Progress (see Section~\ref{evalmetrics}). While the focus of our work is not to train a \questioner, we ensure our \navigator is equipped with the ability to identify \textit{when} to ask questions. This leads to our proposed \textbf{general game-play mode}, wherein the aforementioned description of a regular game-play mode episode still holds but the hard-coded heuristic of asking questions every 4\textsuperscript{th} time-step is eliminated, i.e., the \navigator decides when a question must be asked to continue game-play.

\subsection{\oscar}
The \oscar pre-training strategy~\citep{li2020oscar} for cross-modal Transformers uses object tags detected in images as \textit{anchor points} to ease the learning of semantic alignments between images and text. The input is represented as Word-Tag-Image $(\vw, \vq, \vv)$, where $\vw$ and $\vq$ are the sequence of word embeddings of the text and object tags respectively, and $\vv$ is the sequence of region features of the image. To generate $\vv$, Faster R-CNN~\citep{ren2015faster} is used to extract visual semantics of each region as $(v', z)$ where  $v' \in \sR^P$ ($P = 2048$) is the region feature, $z \in \sR^6$ is the region position represented by the coordinates of the top-right and bottom-left corners and the height \& width. $v'$ and $z$ are concatenated to form a position-sensitive region feature, which is further transformed into $v$ using a projection layer such that $v$ has the same dimension as the input token embeddings. It is then pre-trained with a Masked Token Loss (MTL) and a Contrastive Loss (CL).

\begin{align*}
    \gL_{Pre-training} &= \gL_{MTL} + \gL_{CL} \\
    &= -\E_{(\vv, \vh) \sim \gD} \log p(h_i | \vh_{\backslash i}, \vv) \\
    & -\E_{(\vh', \vw) \sim \gD} \log p(y | f(\vh', \vw))
\end{align*}

The MTL is akin to that in BERT~\citep{devlin2019bert}, masking the input tokens $(\vw, \vq)$ with a probability of 15\% and predicting them. The CL is computed by polluting the object tags $\vq$ with a probability of 50\% with randomly chosen object tags from the dataset, and a feed-forward layer on top of $\clsToken$ predicts whether the input contains the original image representation or a polluted one. In the previous equation, $\vh = [\vw, \vq]$, $\vh' = [\vq, \vv]$, $\vh_{\backslash i}$ are the surrounding tokens of masked token $h_i$, $f(.)$ denotes the binary classifier where $y = 0$ if the object tags are polluted and 1 otherwise, and $\gD$ is the dataset. \oscar uses a collection of popular image-text datasets for pre-training, including but not limited to Conceptual Captions~\citep{sharma2018conceptual}, MS-COCO~\citep{lin2014microsoft}, Flickr30K~\citep{flickr30k} and GQA~\citep{hudson2019gqa}. Such datasets typically have images of objects taken from perfect angles whereas a navigating agent will see objects from different vantage points, which also motivates augmenting \oscar and performing an additional phase of navigation-specific pre-training.

\section{Approach}

\begin{figure*}[h]
\begin{center}
  \includegraphics[width=\linewidth]{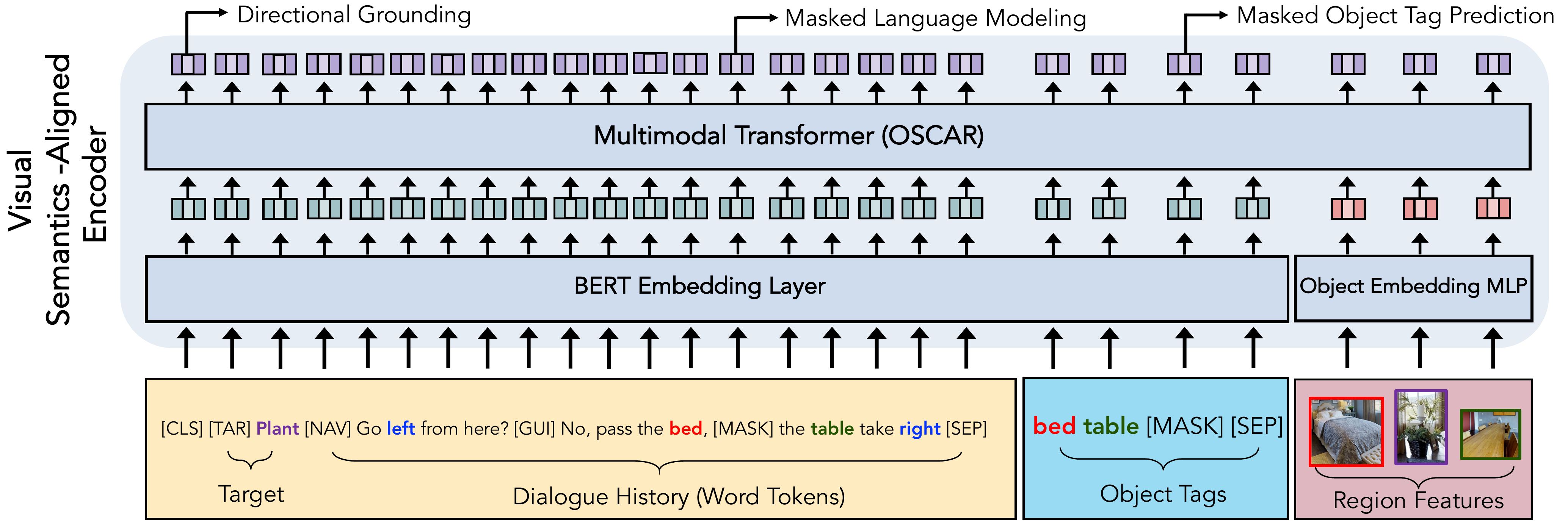}
\end{center}
   \caption{\visitron's Encoder Architecture and Semantics-Aligned Navigation Pre-Training Tasks}
\label{fig:oscar}

\end{figure*}

The policy for NDH (and VLN) can be decomposed into an encoder-decoder setup,  $\pi_{\vtheta} = f_{\vtheta_{E}} \circ f_{\vtheta_{D}}$:

\begin{itemize}
    \item A vision-language encoder $f_{\vtheta_E}: \{\vs_{1:t}, \vx \} \rightarrow \vz_t$, where $\vs_{1:t}$ are visual states, $\vx$ is the \dia history (or instructions for VLN) and $\vz_t$ is the joint latent representation at time step $t$.
    \item An action decoder $f_{\vtheta_D}: \{ \vs_t, \vz_t, \va_{t-1} \}  \rightarrow \va_t$, where $\va_t$ is the next action.
\end{itemize}

We model $\pi_{\vtheta}$ by \visitron, a visio-linguistic Transformer-based model. \visitron's encoder is structurally similar to \oscar's Transformer~\citep{li2020oscar}. This is by design to enable easy transfer of visual semantics-aligned representations learned from disembodied image-text data. We make navigation-specific modifications to \oscar, but they are all structured as augmentations of modules instead of removal of network components, thus enabling us to use the pre-trained weights of \oscar's Transformer to initialize large portions of our encoder. The augmentations are described in Section~\ref{visitron-pretraining}. As with \oscar, the input to \visitron's encoder is represented as Word-Tag-Image $(\vw, \vq, \vv)$, where $\vw$ and $\vq$ are the sequence of word embeddings of the text and object tags respectively, and $\vv$ is the sequence of region features of the image. We represent the panorama in 36 views, extract Faster R-CNN~\citep{ren2015faster} region features $r'$ from each view and add positional vector $p$, $r=(r', p)$. To incorporate 3D direction, we add direction embedding $d$ to the region features, $v = r + d$. $d$ is a 128-dimensional orientation vector represented by repeating $[\sin \phi; \cos \phi; \sin \omega; \cos \omega]$ 32 times where $\phi$ and $\omega$ are heading and elevation poses. In addition to the standard $\clsToken$ and $\sepToken$, we also use $\tarToken$, $\navToken$, $\guideToken$ as delimiter tokens for the initial target hint, \navigator's questions and the \guide's answers respectively. While this input structure is \dia-specific, it is amenable to instructions-based datasets for multi-tasking.

\subsection{\visitronBold Pre-Training}\label{visitron-pretraining}
We adopt a two-stage pre-training strategy, initializing \visitron's encoder with weights from \oscar to begin with web-scale disembodied visio-linguistic representations, followed by facilitating a domain shift to navigation and actions by pre-training on navigation data. For each navigation trajectory, we extract $(\vw, \vq, \vv, \va)$ tuples where $\vw$ is the \dia history/instruction, $\vq$ is the sequence of object tags from the current panorama, $\vv$ is the sequence of region features and $\va$ is the direction in the 360\textdegree\xspace panoramic space where the next node in the trajectory is located~\citep{fried2018speaker}. 
The pre-training objectives are:

\begin{enumerate}
  \item{
    \textbf{Masked Language Modeling:} Input word tokens are replaced with $\maskToken$ with 15\% probability and the masked token $x_i$ is predicted conditioned on surrounding tokens $x_{\backslash i}$.
  }
  \item{
    \textbf{Masked Object Tag Prediction:} Object tags are replaced with $\maskToken$ with 15\% probability. A feed-forward head on top of $\maskToken$ is used to predict the tag from a distribution over Faster R-CNN semantic classes. This provides more fine-grained object supervision unlike \oscar's global masked token loss for tokens in both object tags and text, since this computes a distribution over the object detector's semantic classes instead of over the entire input vocabulary.
  }
  \item{
    \textbf{Directional Grounding:} $\clsToken$ hidden state goes into a feed-forward head to predict $\va$.
  }
\end{enumerate}

Figure~\ref{fig:oscar} illustrates \visitron's encoder architecture and the pre-training objectives we use, with an extracted tuple from a sample NDH instance.

\subsection{\visitronBold Fine-Tuning}\label{visitron-finetuning}
\begin{figure}[t]
\begin{center}
\includegraphics[width=1.0\linewidth]{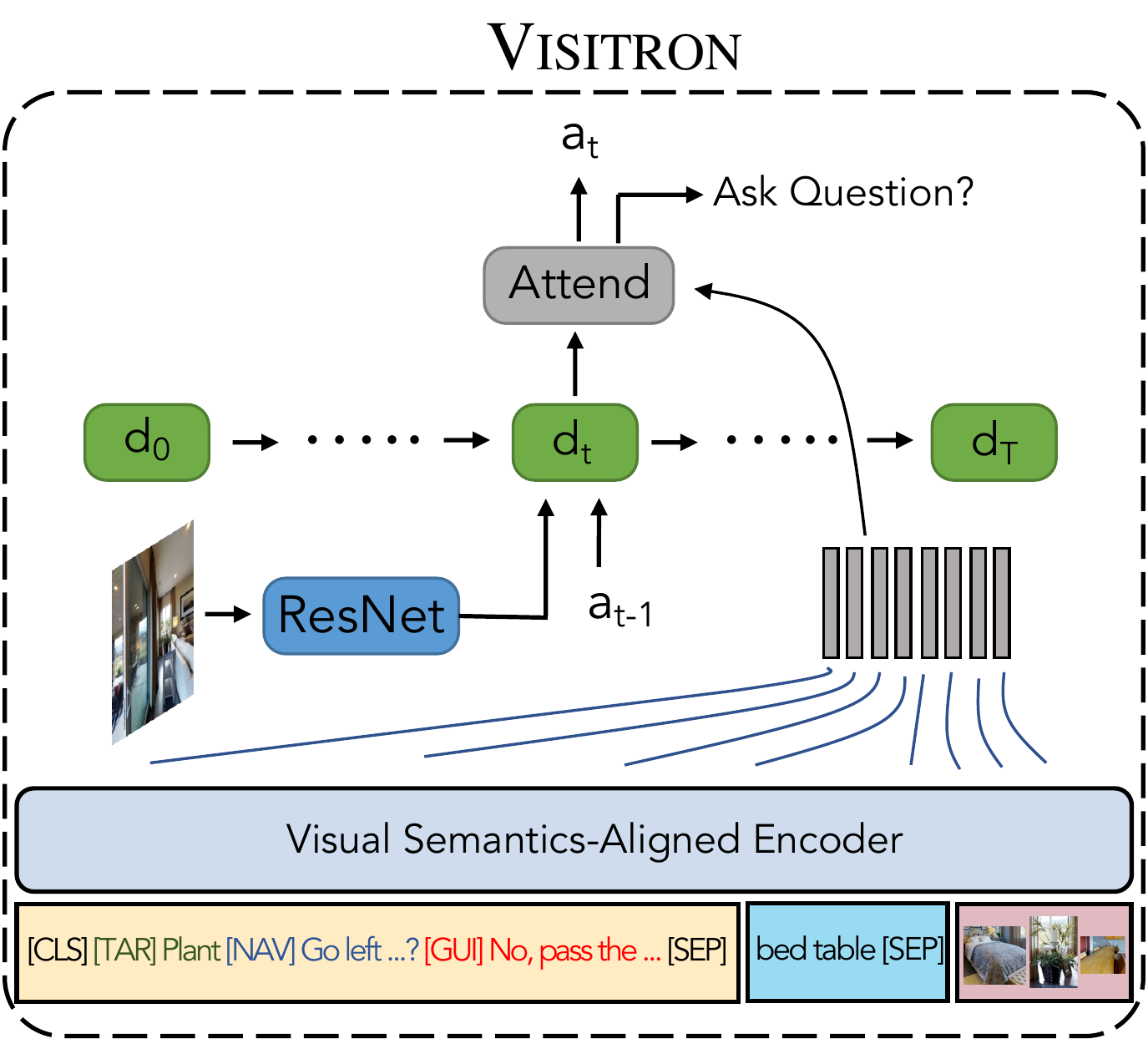}
\end{center}
   \caption{\navigator predicts navigation actions, given \dia history and visual observations. The same stack decides when to ask the \guide a question. A similar setup can be used for question generation.}
\label{fig:navigator}
\vspace{-10px}
\end{figure}

After pre-training the encoder, we leverage it with an attention-based Long Short-Term Memory (LSTM) action decoder~\citep{hochreiter1997long}, as shown in Figure~\ref{fig:navigator}.
At time-step $t$, the decoder (cell state $d_t$) takes the previous action $\va_{t-1}$, the panoramic ResNet features extracted from the current location/state and decodes the next action $\va_t$, while attending to the \visitron encoder's cross-modal representation of its input. After this LSTM is fine-tuned, the same stack is frozen and a randomly initialized two-layer feed-forward head is added and trained with a binary cross-entropy loss to learn to classify when to ask a question. The supervision for this head comes from the elongated CVDN instances defined in Section~\ref{CVDN}, with time-steps when a question was asked serving as positive labels and the remaining time-steps during which navigation occurs serving as negative labels. Note that as described in Section~\ref{VLN}, the decoder's actions can belong in either the panoramic space or the low-level visuomotor space~\citep{fried2018speaker}, leading to independent formulations for \textit{viewpoint selection} and \textit{turn-based action prediction}.

\setlength{\tabcolsep}{3pt}
\begin{table*}[t]
\caption{Pre-Training Ablations (Fine-Tuning and Evaluating on NDH)}
\label{tab:ablation}
\centering
\resizebox{\linewidth}{!}{
\begin{tabular}{lr cc ccc cccc cccc}
\toprule
&& \multicolumn{5}{c}{\bf{Semantics-aligned Pre-Training Curriculum}} & \multicolumn{4}{c}{\multirow{3}{*}{\bf Val Seen}} & \multicolumn{4}{c}{\multirow{3}{*}{\bf Val Unseen}} \\
\cmidrule(l{4pt}r{4pt}){3-7}
&& \multicolumn{2}{c}{Stage 1: Web (\oscar)} & \multicolumn{3}{c}{Stage 2: Navigation} \\
\cmidrule(l{4pt}r{4pt}){3-4}
\cmidrule(l{4pt}r{4pt}){5-7}
\cmidrule(l{4pt}r{4pt}){8-11}
\cmidrule(l{4pt}r{4pt}){12-15}
&\texttt{\#} & \shortstack{Contrastive+\\Masked LM} & \shortstack{Object\\Tags} & \shortstack{Masked\\LM} & \shortstack{Masked Object\\Tag Prediction} & \shortstack{Directional\\Grounding} & \GP (m) $\uparrow$ & \SPL (\%) $\uparrow$ & \SR (\%) $\uparrow$ & \nDTW (\%) $\uparrow$ & \GP (m) $\uparrow$ & \SPL (\%) $\uparrow$ & \SR (\%) $\uparrow$ & \nDTW (\%) $\uparrow$ \\
\toprule
\multirow{7}{*}{\rotatebox{90}{\visitron}} & 1 & \multicolumn{5}{c}{(No pre-training and no object tags)} & 4.76 & 36.56 & 46.07 & 30.97 & 2.09 & 9.96 & 22.49 & 6.50 \\
\cmidrule(l{4pt}r{4pt}){3-15}
 & 2 & \cmark &&&&& 4.82 & 50.73 & 58.11 & 47.34 & 2.67 & \textbf{24.88} & \textbf{34.29} & 24.21 \\
& 3 & \cmark & \cmark &&&& 4.38 & 45.15 & 52.09 & 41.14 & 2.30 & 13.03 & 24.81 & 8.63 \\
& 4 & \cmark & \cmark & \cmark &&& 5.09 & 25.92 & 41.10 & 17.91 & 1.90 & 11.27 & 23.48 & 5.62 \\
& 5 & \cmark & \cmark& \cmark & \cmark && 4.83 & 48.22 & 56.02 & 47.01 & 2.70 & 24.04 & 32.86 & 23.46 \\

\cmidrule(l{4pt}r{4pt}){3-15}
& 6 & \cmark & \cmark & \cmark & \cmark & \cmark & \textbf{5.34} & \textbf{55.16} & \textbf{61.78} & \textbf{54.83} & \textbf{2.71} & 24.56 & 32.52 & \textbf{24.51} \\
\bottomrule
\end{tabular}
}
\end{table*}
\setlength{\tabcolsep}{1.4pt}

\section{Experiments}
In this section, we first describe the evaluation metrics we adopt. We then describe and discuss our experimental observations from performing ablations during \visitron pre-training and fine-tuning respectively. We present our observations for question-asking classification for CVDN, establishing a strong baseline for future models. We finally present and discuss our observations from submitting our model checkpoints to the \textit{static} EvalAI leaderboard for CVDN.

\subsection{Evaluation Metrics}
\label{evalmetrics}
We evaluate \visitron's ability to navigate to the goal with the following metrics:

\begin{itemize}
    \item \textit{Goal Progress} (\GP) measures the difference between the distance from the start position to the final goal and the distance from the end position to the final goal. It is used to determine how much progress in meters the agent has made towards the final goal.
    \item \textit{Success weighted by (Normalized Inverse) Path Length} (\SPL) introduced by \newcite{anderson2018evaluation} provides a measure of success normalized by the ratio between the length of the shortest path and the selected path.
    \item \textit{Success Rate} (\SR) measures the success of an episode. If the agent stops within 3 meters of the goal, it is considered a success.
    \item \textit{Normalized Dynamic Time Warping} (\nDTW) introduced by \newcite{ilharco2019general} helps measure a navigator agent's fidelity to the \dia history/instruction by softly penalizing deviations from the reference path.
\end{itemize}

We evaluate the question-asking classification head by computing \textit{accuracy} and \textit{balanced accuracy}~\citep{10.1109/ICPR.2010.764}. The latter accounts for the natural class imbalance of more navigation time-steps than question-asking time-steps expected in \dia-based navigation by computing the average of recall obtained on each class.

\setlength{\tabcolsep}{3pt}
\begin{table*}[t]
\caption{Fine-Tuning Ablations}
\label{tab:dataset_multitask}
\centering
\resizebox{\linewidth}{!}{
\begin{tabular}{lr cc cccc cccc}
\toprule
&& \multirow{2}{*}{\shortstack{\bf Action\\\bf Space}} & \multirow{2}{*}{\shortstack{\bf Multi-Task\\\bf Fine-Tuning\\ NDH+}} &  \multicolumn{4}{c}{\bf Val Seen} & \multicolumn{4}{c}{\bf Val Unseen} \\
\cmidrule(l{4pt}r{4pt}){5-8}
\cmidrule(l{4pt}r{4pt}){9-12}
&\texttt{\#} &&& \GP (m) $\uparrow$ & \SPL (\%) $\uparrow$ & \SR (\%) $\uparrow$ & \nDTW (\%) $\uparrow$ & \GP (m) $\uparrow$ & \SPL (\%) $\uparrow$ & \SR (\%) $\uparrow$ & \nDTW (\%) $\uparrow$ \\
\cmidrule(l{4pt}r{4pt}){2-12}
\multirow{4}{*}{\rotatebox{90}{\visitron}} & 1 & \multirow{2}{*}{\shortstack{Turn-based\\Action Prediction}} & \xmark &  1.15 & 9.66 & 11.78 & 26.86 & 1.60 & 13.02 & 14.77 & 29.28  \\
& 2 && \cmark (RxR) & 1.50 & 12.30 & 15.18 & 19.95 & 0.97 & 11.52 & 15.44 & 20.49  \\
\cmidrule(l{4pt}r{4pt}){2-12}
\morecmidrules
\cmidrule(l{4pt}r{4pt}){2-12}
& 3 & \multirow{2}{*}{\shortstack{Viewpoint\\Selection}} & \xmark  & 5.34 & 55.16 & 61.78 & 54.83 & 2.71 & 24.56 & 32.52 & 24.51  \\

& 4 && \cmark (RxR) & 5.11 & 12.33 & 25.65 & 4.66 & 3.25 & 10.74 & 27.34 & 3.78  \\

\bottomrule
\end{tabular}
}
\end{table*}
\setlength{\tabcolsep}{1.4pt}

\subsection{Pre-Training Ablations}

Using NDH and R2R trajectories, we pre-train \visitron as described in Section~\ref{visitron-pretraining}. We begin experimenting with cumulative addition of each pre-training stage and objective to obtain an ablative understanding of their effect on the downstream NDH task. Results are shown in Table \ref{tab:ablation}. 
We see that \textbf{our pre-training strategy helps}: the best performance on Val Seen (as measured by all metrics) is obtained when using all pre-training stages and objectives. We also see that Goal Progress (GP) is highest on Val Unseen in this setting (an absolute increase of 0.62 relative to no pre-training).
Rows 3-4 demonstrate the efficacy of our second-stage masked language modeling (MLM) task, helping improve Val Seen GP from 4.38 to 5.09. Rows 4-5 demonstrate the efficacy of our newly introduced masked object tag prediction task as a means towards experience-driven concepts and semantics identification and acquisition, with significant increases in all metrics across both validation seen and unseen splits. Rows 5-6 show that our directional grounding task for pre-training the encoder plays a particularly important role: the increase in both GP and nDTW suggest that this task improves \visitron's ability to navigate closer to the goal while ensuring that \dia fidelity is maintained in the process by aligning encoder representations in the direction along the reference path.

\subsection{Fine-Tuning Ablations}\label{finetune-ablations}
Next, we perform ablations during fine-tuning, leveraging all objectives from Table \ref{tab:ablation} since our previous analysis demonstrated their effectiveness. For VLN agents, it has been shown that viewpoint selection in the panoramic space is a better formulation than turn-based action prediction in the low-level visuomotor space~\citep{fried2018speaker}. However, it is not immediately obvious or known whether this can be extrapolated to \dia-based navigation as in CVDN. So we experiment with both formulations for our \navigator. Given the sparsity of NDH instances ($\sim$ 4k) for fine-tuning, we also study if multi-task fine-tuning with the RxR dataset helps boost performance. Table \ref{tab:dataset_multitask} presents the fine-tuning ablation results.
Row 1 and 3 demonstrate that \textbf{panoramic viewpoint selection is a better formulation than turn-based action prediction for CVDN}, with all metrics increasing significantly when switching to viewpoint selection. Further, we see in rows 3 and 4 that \textbf{multi-task fine-tuning leads to better CVDN generalization}, with Val Unseen GP increasing from 2.71 to 3.25 when multi-tasking with viewpoint selection. However, we see this increase in GP occurs alongside a decrease in nDTW, SPL and SR. This decrease can be attributed to the fact that the RxR dataset has very long trajectories, which prime the model to take long paths to the final CVDN goal (which GP cares about), well-beyond the next 5 \guide steps in the NDH instance that nDTW, SPL and SR evaluate against.

\subsection{Question-Asking Classification and Leaderboard Evaluation}

We pick the \visitron model checkpoint with the highest GP in Table \ref{tab:dataset_multitask} (row 4), and perform imitation learning of the question-asking classification head as described in Section~\ref{visitron-finetuning}. We evaluate the classification head by creating elongated CVDN instances from the validation sets as described in Section~\ref{CVDN}, akin to how supervision was provided during training: time-steps when a question was asked serve as positive instances and the remaining time-steps during which navigation occurs serve as negative instances. As seen in Table \ref{tab:classifier}, our approach to identifying when to ask questions vs.~when to navigate establishes a strong baseline for future work on identifying when to ask questions with CVDN, as measured by accuracy and balanced accuracy on Val Unseen. It is important to note that our design choice of adding and training a separate head for this task while keeping the navigator stack frozen ensures that there is no direct impact on navigation performance itself. This is unlike approaches that perform direct navigation action space augmentation with a special action for question-asking, where navigation actions themselves are affected by the presence of an additional competing variable for shared total probability mass.

\setlength{\tabcolsep}{3pt}
\begin{table}[h!]
\caption{Question-Asking Classification Performance}
\label{tab:classifier}
\begin{center}
\small
\begin{tabular}{l cc}
\toprule
\textbf{Metric (\%)} & \textbf{Val Seen} & \textbf{Val Unseen} \\
\midrule
Accuracy & 68.05 & 67.87 \\
Balanced Accuracy & 63.33 & 61.09 \\
\bottomrule
\end{tabular}
\end{center}
\end{table}
\setlength{\tabcolsep}{1.4pt}

We submitted this model checkpoint to the CVDN leaderboard aimed at the \textit{static} NDH task. We observe in Table \ref{tab:leaderboard} that this model checkpoint's performance is competitive with state-of-the-art models with a hidden test GP of 3.11. However, the low hidden test SPL of 12 indicates the impact that multi-task fine-tuning with long RxR paths had on this checkpoint's ability to take short paths to the goal, like we discussed earlier in Section~\ref{finetune-ablations}. Given this expected decrease in SPL when utilizing such long trajectories, we also created a model checkpoint by multi-task fine-tuning \visitron on NDH, R2R and R4R. We observe that this model checkpoint obtains state-of-the-art SPL of 25 alongside an associated decrease in GP to 2.40.

\setlength{\tabcolsep}{3pt}
\begin{table}[h!]
\caption{NDH Hidden Test Set Performance}
\label{tab:leaderboard}
\begin{center}
\resizebox{\columnwidth}{!}{
\begin{tabular}{rl cc}
\toprule
\texttt{\#} &\textbf{Method} & \GP (m) $\uparrow$ & \SPL (\%) $\uparrow$ \\
\midrule
1 & MT-RCM + EnvAg~\citep{wang2020environment} & 3.91 & 17 \\
2 & BabyWalk~\citep{zhu2020babywalk} & 3.65 & 11 \\
3 & \visitronBold & 3.11 & 12 \\
4 & Cross-modal Memory Network~\citep{zhu2020visioncmm} & 2.95 & 14 \\
5 & PREVALENT~\citep{hao2020towards} & 2.44 & 24 \\
6 & \visitronBold (Best SPL) & 2.40 & \textbf{25} \\
\bottomrule
\end{tabular}
}
\end{center}
\end{table}
\setlength{\tabcolsep}{1.4pt}

\vspace{-0.1in}
\section{Related Work} \label{relwork}

Vision-and-language pre-training~\citep{tan2019learning, lu2019vilbert, sun2019videobert, chen2020uniter, zhou2020unified} has grown to become a popular area of research, primarily aimed at solving downstream tasks such as image captioning, visual question answering and image retrieval. This line of work typically involves learning cross-modal representations using self-supervised objectives with a co-attention Transformer that fuses the two modalities represented by input token embeddings and visual region features, where the latter is typically sourced from Faster R-CNN~\citep{ren2015faster}.

Research in vision-and-language navigation (VLN) has also seen tremendous progress~\citep{fried2018speaker, ke2019tactical, anderson2019chasing, tan2019learning, zhu2020vision} since the advent of the Room-to-Room (R2R) dataset~\citep{anderson2018vision} based on Matterport3D~\citep{Matterport3D}, with scope for further advances only increasing with the recent release of the much larger, densely annotated and multilingual Room-across-Room (RxR) dataset~\citep{ku2020room}. As an extension to VLN, the recent Cooperative Vision-and-Dialog Navigation (CVDN) dataset~\citep{thomason2020vision} allows for training interactive navigator and guide agents. The dominant focus of research with CVDN so far has been the Navigation from Dialog History (NDH) task introduced with CVDN, which is equivalent to treating the \dia history as a VLN-style fixed instruction. The NDH formulation allows for easy transfer and multi-task learning~\citep{hao2020towards, wang2020environment, zhang2020diagnosing} with VLN.
However, state-of-the-art VLN models such as VLN-BERT~\citep{majumdar2020improving} rely on the fully-observable setting when framing the task as \textit{ahead-of-time} path selection, which is fundamentally at odds with the need for \dia in CVDN: \dia is aimed at enabling the navigating agent to succeed \textit{while} it makes navigation decisions and decides it needs assistance. The recent Recursive Mental Model (\rmm)~\citep{roman2020rmm} for CVDN attempts to address this by introducing a simulated dialogue game-play mode, where a trained navigator is fine-tuned jointly with a pre-trained guide and evaluated in this mode. However, the \rmm navigator does not dynamically ask questions, instead relying on a data-driven heuristic of asking questions after every 4th navigation time-step. \visitron's design naturally leads to a generalization of this game-play mode which eliminates the aforementioned heuristic.

Our work is similar to recent work~\citep{hao2020towards} on leveraging pre-trained cross-modal representations for the NDH task. However, our work takes on added goals of learning when to ask questions and associative learning of visio-linguistic concepts and semantics to ensure they can be identified and acquired when interacting in new environments, which are key requirements for full cooperative vision-and-\dia navigation.

\section{Conclusion and Future Work}

We presented \visitron, a Transformer-based navigator designed to identify and acquire visio-linguistic concepts and semantics and make decisions, all key traits for interactive navigation inherent to CVDN. We demonstrated the efficacy of our approach via experiments and ablations. We proposed generalizing the game-play regime introduced with \rmm~\citep{roman2020rmm} to enable interactive fine-tuning and evaluation of \visitron-like models with pre-trained guides. The trade-off between GP and SPL in \dia-based navigation, Sim-to-Real transfer~\citep{anderson2021sim} and robustness in \dia-based navigation in presence of speech recognition errors~\citep{gopalakrishnan2020neural} are all important problems that merit detailed investigation in future work.



\section{Societal Impact}

The primary dataset of interest for our work on interactive navigation in photo-realistic indoor environments: Cooperative Vision-and-Dialog Navigation (CVDN), is an English-only dataset. We also multi-task with several other datasets, namely R2R, R4R and RxR, but RxR is the only multilingual dataset and covers English, Hindi and Telugu. Due to CVDN being English-only, we utilized the English-portion of the RxR data during multi-task fine-tuning. There are over 6500 known languages spoken in the world today and vision-and-dialog navigation research could, in principle, be deployed in every home in the world, but due to current data limitations, it can only be deployed in English-speaking homes. Our modeling methods should transfer to other languages given sufficient volume of data, but ensuring that might not be possible for low-resource or endangered languages. \visitron may benefit from new training schemes and modeling improvements to account for such scenarios. When deployed in real homes, speech would be the primary modality for most humans to interact with such robots. While speech recognition research has advanced considerably, ensuring accurate speech recognition across various speaker populations and accents is still challenging. Errors in speech recognition could impact \visitron's ability to navigate accurately, so making \visitron robust to speech recognition errors will be necessary, potentially via augmentation of the language component of its training data with synthetic and actual speech recognition errors~\citep{gopalakrishnan2020neural}.

During navigation, \visitron needs access to neighboring viewpoints to select from. Each environment in CVDN contains an underlying navigation graph which provides this information, which might not be the case in real unseen environments. In its absence, additional modules can be added that generate a local navigation graph based on the surroundings~\citep{anderson2021sim}. Datasets in the vision-and-language navigation space such as R2R and CVDN typically consider the environment to be static. Obstacle avoidance methods need to be added to models built using these datasets to avoid hazardous collisions in a dynamic environment, such as with moving humans and pets.

Large language models are known to have a high carbon footprint associated with training them~\citep{strubell2019energy}. \visitron is about the same size as BERT~\citep{devlin2019bert}, which is now ubiquitously used in both academic and industrial settings and can be trained reasonably fast. The carbon footprint of this work was maintained within permissible limits by using a maximum of 8 Tesla V100 GPUs for training.

\section*{Acknowledgments}

Many thanks to Jesse Thomason and Aishwarya Padmakumar for useful technical discussions and actionable feedback on multiple versions of this paper. We would also like to thank the anonymous reviewers for their service and useful feedback.

\bibliography{anthology,custom}
\bibliographystyle{acl_template/acl_natbib}

\end{document}